# An Expert System Approach for determine the stage of UiTM Perlis Palapes Cadet Performance and Ranking Selection

Tajul Rosli B. Razak

Faculty of Computer and Mathematical Sciences
Universiti Teknologi Mara (Perlis)
tajulrosli@perlis.uitm.edu.my

*Abstract:* The Palapes cadets are one of the uniform organizations in UiTM Perlis for extra-curricular activities. The Palapes cadets arrange their organization in a hierarchy according to grade. Senior Uniform Officer (SUO) is the highest rank, followed by Junior Uniform Officer (JUO), sergeant, corporal, lance corporal, and lastly, cadet officer, which is the lowest rank. The Palapes organization has several methods to measure performance toward promotion to higher rank, whether individual performance or in a group. Cadets are selected for promotion based on demonstrated leadership abilities, acquired skills, physical fitness, and comprehension of information as measured through standardized testing. However, this method is too complicated when manually assessed by a trainer or coach. Therefore, this study will propose an expert system, which is one of the artificial intelligence techniques that can recognize the readiness and progression of a Palapes cadet. Readiness is the degree to which a cadet is capable of fulfilling duties involving leadership skill and progression is the rate or manner to which a cadet is assigned new leadership or managerial duties. This study also will ensure that all eligible Palapes cadets will receive a full and equitable opportunity to compete for promotion and for the trainers to actively monitor their cadets' performance.

## 1. Palapes Cadet Measurement

In military organizations, transparent and fair appraisal of personnel is essential for decisions pertaining to promotions and operations [1]. Currently, there is a manual process to measure Palapes UiTM cadet performance to increase their rank. The cadet officers in the Palapes organization are abundant. In addition, they all have talent to become leaders or captains in their platoon. Therefore, the process to choose a captain or leader requires time and energy. The Palapes organization uses a manual selection method to determine a very charismatic leader. The following selection methods are currently applied in the Palapes organization. First, they hold oral and paper examinations. The results are compared between the cadet officers to determine which is the best and highest rank. Second, the command tests individuals. Those who have the spirit of leadership will have a bright opportunity to be promoted to a higher rank. Third, physical fitness us tested. Once their fitness reaches the level determined appropriate by Palapes, that person can be promoted to a higher rank. From overall observation, we can see these processes to choose a leader are too complicated.

To provide an easier way to measure Palapes cadet performance for promotion to higher rank, this study has developed an expert system. The system will correctly choose a person who can be promoted to higher rank from all of the information obtained from the cadet officer who is an expert in this study. The system will match their capability in leadership with the apposite rank in the Palapes organization and will overwrite the current system to more easier organize the process and reduce the complexity.

## 2. Objective

The main purpose of this study is to implement an expert system that facilitates the process of measuring Palapes cadet performance and identify which cadets are best fit for promotion or the next higher rank. The other objectives are as follows:
- To ensure that all eligible cadets receive full and equitable opportunity to compete for promotion
- To store and retrieve cadet grade information in an easier and more efficient manner
- To provide a backup for the information related to cadet grading
- To save time, resources, and cost.

## 3. Expert System

The creation of an expert system normally requires specific technical knowledge and concepts of artificial intelligence (AI). Expert systems have provided solutions to multiple problems in companies of all types [2]. However, developing an expert system based on these tools becomes a very difficult task for users without specific training in small and medium-sized companies. A tool that is easy to use but still has enough power to solve problems and can be used by the domain expert makes the technology of expert systems accessible in all types of companies. The basic idea behind an expert system is simply expertise, which is the vast body of task-specific knowledge being transferred from a human to a computer. This knowledge is then stored in the computer and users call upon the computer for specific advice as needed [3]. This study will show how an expert system had been applied to measure Palapes cadet performance to determine who will be promoted to the next higher rank.

### 3.1 Knowledge Acquisition and Knowledge Representation

Some knowledge acquisition methods are available to be considered for expert system development and the form of a direct approach whereby the specialized knowledge interacts directly with the human expert, generally through interviews or questionnaires, obtaining an explanation of the knowledge that the expert applied to solve a particular problem [4]. After the problem has been selected, the knowledge acquisition phase of expert system development is begun using the direct approach. The task now is to have the knowledge that the expert uses to solve the problem displayed in logical fashion so that it can be coded into the computer. We have used the interview method to collect the knowledge from experts. The domain knowledge is then translated into diagrams and a set of rules are then stored in a knowledge base that will be used for the inference engine or network of an expert system.

A knowledge base consists of problem-solving rules, procedures, and intrinsic data relevant to the problem domain [5]. A shell stores the copious knowledge gathered from rules, operations guides, assessment frameworks, experts, historical data, and books regarding the application in the form of rules. It also can be either factual or inferential [6]. The most common definition of knowledge base is human-centered. This highlights the fact that knowledge bases have their roots in the field of artificial intelligence and that they are attempts to understand and initiate human knowledge in computer systems. The four main components of a knowledge base are usually distinguished as a knowledge database, an inference engine, a knowledge engineering tool, and a specific user interface.

Fig. 1 shows the conceptual diagram used to visualize the overall process in this study. The cadet performance will be evaluated through standard testing in a Palapes organization. Then, from that score, it will be determined whether the cadet will be promoted to the next rank based on the stage of performance. Thereby, the standard testing that has been operating under the Palapes organization is show in Table 1.

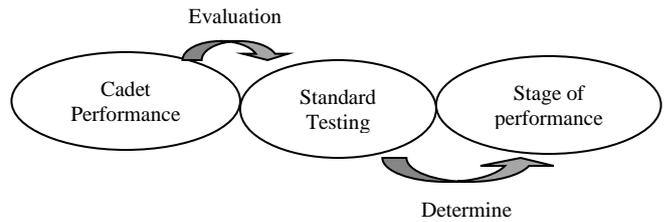

**Figure 1**. Conceptual Diagram

**Table 1.** Standard Testing

| Test | | Percentages (%) |
|---|---|---|
| **Leadership** | | 14 |
| **Theory** | Paper 1 | 12 |
| | Paper 1 | 12 |
| **Military Skills** | Practical | 12 |
| | IMP | 12 |
| | Marching | 6 |
| | Weapons | 6 |
| | Shooting Skill | 4 |
| | War Field | 10 |
| **Sports** | | 3 |
| **Attendance** | | 6 |
| **Coach Observation** | | 3 |
| **TOTAL** | | 100 |

The scores of the standard testing are very important to determine the cadet's chance to be promoted for the next rank. However, if the score is below 50%, the cadets automatically do have no chance to be promoted. For this study, an expert system approach will be applied to see the stage of cadet performance from their standard testing score in order to be promoted to the next higher rank in Palapes cadetin UiTM Perlis.

Fig. 2 is an illustration of the process of determining the stage of cadet performance based on their standard testing score. This process is conducted through the inference network in the expert system. Ranges of scores in standard testing for each stage of cadet performance are shown in Table 2.

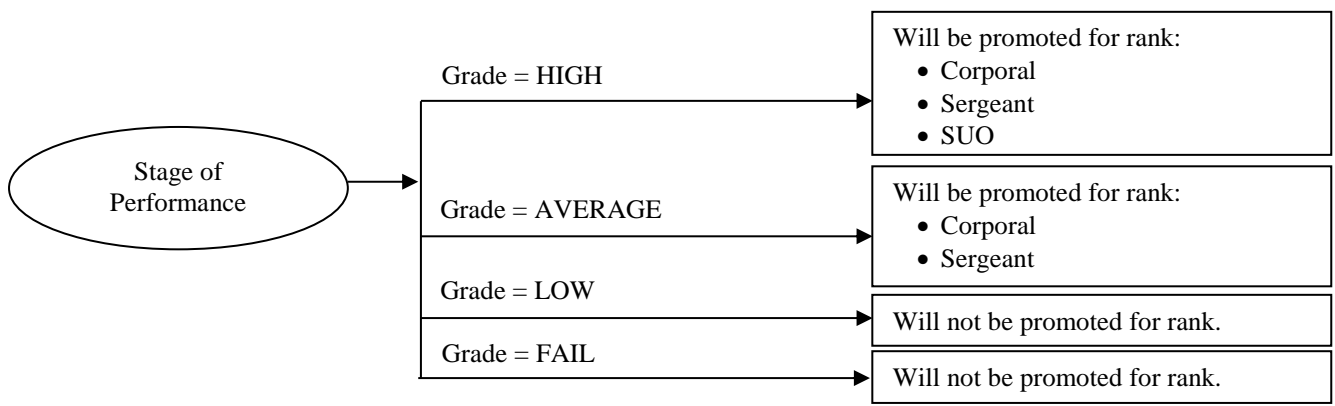

**Table 2.** Stages of Cadet Performance

| **HIGH** | 80 - 100 |
| **AVERAGE** | 60 - 79 |
| LOW | 50 - 59 |
| FAIL | < 50 |

If the standard testing scores are in the range of 80% to 90%, the cadet performance stage is HIGH. While if the standard testing scores are in the range of 60% to 79%, the cadet performance stage is AVERAGE. Standard testing scores in the range of 50% to 59% are LOW. The last stage of cadet performance is FAIL, which is standard testing scores below 50%.

All of this information will be stored in a knowledge base in the expert system. This will be converted to rule-based information before being stored in the knowledge base. An expert system will utilize inference or reasoning in the problem using forward chaining and backward chaining in the explanation. The expert system used the rules in the knowledge base instead of using another data method because it is easy to track and retrieve the solution. The expert system is distinct when compared to others because it can provide an explanation concerning this solution.

| | |
|---|---|
| **IF** | Grade = HIGH (80 - 100 %) |
| **THEN** | Promote for Rank (Corporal, Sergeant, SUO, JUO) |
| **ELSE IF** | Grade = AVERAGE (60 - 79 %) |
| **THEN** | Promote for Rank (Corporal, Sergeant) |
| **ELSE IF** | Grade = LOW (50 - 59 %) |

Fig. 3 shows some of the rule-based data that have been stored in the knowledge base of this expert system. This study is a discussion of the derivation of the cadet performance solution used to promote a cadet to the next rank.

## 4. Result And Discussion

This study is a summary of an expert system that can be applied to measure Palapes cadet performance to select which cadet to promotion to the next rank. Therefore, the next phase is a discussion of the results obtained from this study.

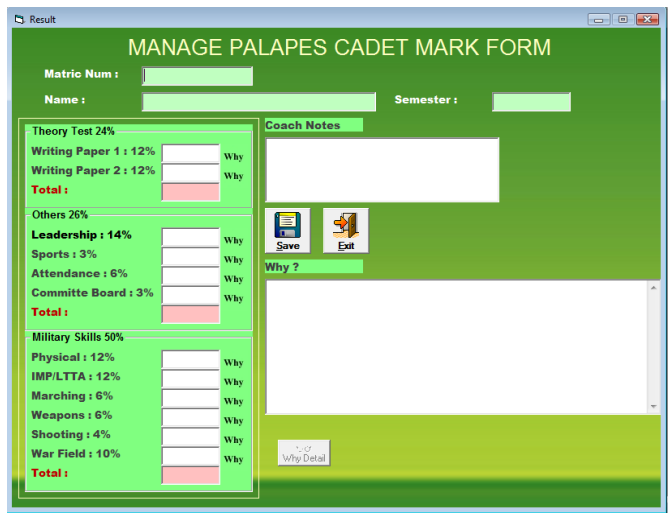

**Figure 4.** Manage Palapes cadet marks form

Fig. 4 shows the interface of this expert system to measure Palapes cadet performance. A cadet officer will use this interface to measure the performance of Palapes cadets based on all of the standard testing values. After all of the standard testing has been completed, the system will calculate and determine the stage of cadet performance as show in Fig 5.

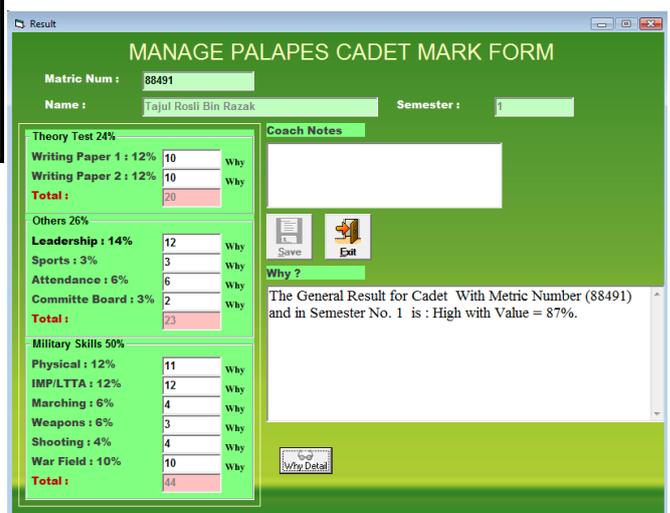

**Figure 5.** The general explanation of the solution given

This system is unique in that it provides an element where a coach or officer makes notes about the Palapes cadet based on observation. If two Palapes cadets are at the same stage of performance (e.g., HIGH), to select one for promotion to the next rank, the officer will examine the coaches' or officer's notes. This can provide an extra bonus for the Palapes cadet.

As show in Fig. 6, this system also provides an explanation concerning the solution if they would like to see how a Palapes cadet obtained a particular result. In this way, the process becomes more reliable.

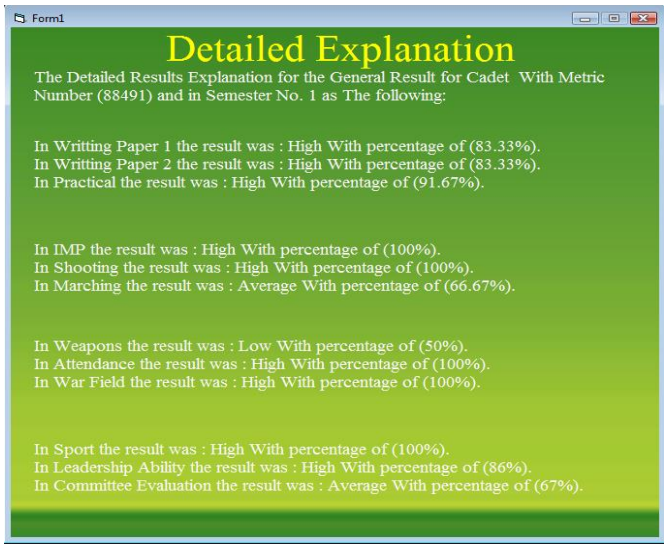

**Figure 6.** The detailed explanation concerning the solution given

## 5. Conclusion

This study shows an expert system developed with the ability to identify cadet levels for decision makers, according to their grades and coach observation, and to integrate the decisions with self-explanatory functions. Cadets' grades will be corroborated with the necessary information. This system mainly is design for UiTM Perlis Palapes but it can be used by any other university's Palapes or defense if they use the same evaluation scheme to promote and grade their cadets. The model also can be used as a reference by other groups to enhance their evaluation system.

## 6. Acknowledgement

The study presented is supported directly and indirectly by the Palapes Uitm Perlis organization to complete this expert system.